\documentclass[letterpaper, 10 pt, conference]{ieeeconf}
\IEEEoverridecommandlockouts                             
\usepackage[utf8]{inputenc}
\usepackage[T1]{fontenc}
\usepackage{graphicx}

\usepackage{subfigure}
\usepackage{color}

\usepackage{float}
\usepackage{lipsum}
\usepackage{stfloats}
\usepackage{multicol}
\usepackage{multirow}
\usepackage{amsmath,bm}
\usepackage{etoolbox}

\usepackage{array}

\usepackage{tabulary}
\usepackage{xcolor}

\usepackage{makecell}

\usepackage{amsfonts}
\definecolor{mypink}{RGB}{232, 166, 181}

\title{\LARGE \bf
Perception Framework through Real-Time Semantic Segmentation and Scene Recognition on a Wearable System for the Visually Impaired
}

\author{Yingzhi Zhang$^{1}$, Haoye Chen$^{1}$, Kailun Yang$^{1}$, Jiaming Zhang$^{1}$ and Rainer Stiefelhagen$^{1}$
\thanks{This work was supported in part by the Federal Ministry of Labor and Social Affairs (BMAS) through the AccessibleMaps project under Grant 01KM151112, in part by the University of Excellence through the ``KIT Future Fields'' project, in part by the Hangzhou SurImage Technology Company Ltd., and in part by the Hangzhou KrVision Technology Company Ltd. (krvision.cn).}
\thanks{$^{1}$The authors are with Institute for Anthropomatics and Robotics, Karlsruhe Institute of Technology, 76131 Karlsruhe, Germany (correspondence: kailun.yang@kit.edu).}
}

\begin{document}

\maketitle
\thispagestyle{empty}
\pagestyle{empty}

\begin{abstract}

As the scene information, including objectness and scene type, are important for people with visual impairment, in this work we present a multi-task efficient perception system for the scene parsing and recognition tasks. Building on the compact ResNet backbone, our designed network architecture has two paths with shared parameters. In the structure, the semantic segmentation path integrates fast attention, with the aim of harvesting long-range contextual information in an efficient manner. Simultaneously, the scene recognition path attains the scene type inference by passing the semantic features into semantic-driven attention networks and combining the semantic extracted representations with the RGB extracted representations through a gated attention module. In the experiments, we have verified the systems' accuracy and efficiency on both public datasets and real-world scenes. This system runs on a wearable belt with an Intel RealSense LiDAR camera and an Nvidia Jetson AGX Xavier processor, which can accompany visually impaired people and provide assistive scene information in their navigation tasks.
\end{abstract}

\section{Introduction}

According to the World Health Organization, globally, the number of people with visual impairment is estimated to be 285 million, of whom 39 million are blind~\cite{bourne2017magnitude}.
Low vision or blindness makes people with visual impairment suffer on the daily basis, let alone in unfamiliar situations.
When visually impaired individuals are in an unknown environment, one of the most vital information for them is their surrounding information.
Visually impaired people desire to confirm whether they are in the correct place,
knowing the specific scene where they are,
for example, in a canteen or on the street.
At the same time, they are concerned about the objects that are ahead of them. Within this context, this work aims to integrate scene recognition and semantic segmentation in a single perception system to help visually impaired people gain more environmental information with a real-time inference speed for fast responses.

Many research efforts pay attention to efficient semantic segmentation for real-time scenarios.
In the former deep learning methods, architectures such as FCN~\cite{long2015fully} and SegNet~\cite{badrinarayanan2017segnet} learn strong features based on various combinations of CNN methods. The early architectures of semantic segmentation generally use slow and complicated VGG~\cite{simonyan2014very} as the basic structure, making the real-time application on wearable devices less likely.
Differing from conventional methods, pieces of research works~\cite{yang2018unifying}\cite{nekrasov2018lightweight}\cite{sun2020real} proposed lightweight networks to decrease floating-point operations.
However, with the finite obstacles that frequently appear in various scenes, visually impaired people cannot accurately grasp the specific scene type. 
A second branch or works~\cite{wang2017weakly}\cite{cheng2018scene} achieved the scene recognition task through concerning objects the scene include or the localization information.

The above works perform well on either semantic segmentation task or scene recognition task.
This work, instead, aims to construct a multi-task system applicable with real-time operating speed, which provide the surrounding objects and the scene information, simultaneously.
Our work is concerned about the lightweight network to process scenarios captured by sensors for more efficient computing and processing capabilities.
The model is deployed on a wearable system composed of a Realsense L515 and an Nvidia Jetson AGX Xavier processor so that the wearable system is able to accompany visually impaired people and assist them with acoustic feedback, e.g., by combining the system with bone-conduction earphones on the wearable glasses~\cite{chen2020can}.
A comprehensive set of experiments on both public datasets and real-world scenes captured by our wearable system, demonstrates the effectiveness of the presented universal perception framework.

\section{Related Work}

\subsection{Visual Assistive System}
Assistive technology is introduced to help visually impaired people in their daily lives.
Various wearable devices are equipped with sensors to capture the surrounding scene, such as RGB cameras, supporting color sensing and image processing. User interfaces in the visual assistant system are always given through audio feedbacks~\cite{rodrguez2012assistingTV} and sonified information~\cite{hu2020comparative}.
Until now, vision-based assistive systems have been employing deep learning architecture to improve perception tasks~\cite{mao2020can}.
The assistive system proposed by~\cite{yimin2019Deep} learns from RGBD data and predicts semantic maps to support the obstacle avoidance task.
\cite{elmannai2018highly} integrated sensor-based, computer vision-based, and fuzzy logic techniques to detect objects for collision avoidance.
A CNN-based framework DEEP-SEE~\cite{tapu2017deepsee}, integrated into a novel assistive device, was designed to recognize objects in the outdoor environment.
In~\cite{lin2018krnet}, a kinetic real-time CNN was customized for the recognition of road barriers to support navigation assistance for the visually impaired, which are usually set at the gate of a residential area or working area.
The wearable system~\cite{chen2020can} with a pair of smart glasses informs the visually impaired people based on semantic segmentation and 3D reconstruction. Differently, our work puts the focus on the multi-task model and satisfying the real-time requirement simultaneously. Our deep learning system produces the exact and flexible assistive information described with the surrounding object information and the scene type.

\subsection{Efficient Semantic Segmentation}
In recent years, many research efforts of real-time semantic segmentation have been devoted to designing compact architectures, aiming to reduce the computational load.
Most approaches take into account both efficiency and accuracy.
According to the analysis, the image resolution affects the speed of computation. Most approaches are down-sampling the input, down-sampling the features, or performing model compression to accelerate their models.
ICNet~\cite{zhao2018icnet} takes advantage of the efficient processing and high inference quality by fusing the low- and high-resolution pictures.
Group convolution in AlexNet~\cite{krizhevsky2012imagenet} and ResNeXt~\cite{xie2017aggregated} improve the effectiveness and reduce the computational cost. ShuffleNet~\cite{zhang2018shufflenet} makes use of channel shuffle to solve the problem of information passing across convolutional groups. MobileNet~\cite{howard2017mobilenets} introduces the decomposable depthwise separable convolution and reduces parameters. BiSeNet~\cite{yu2018bisenet} makes full use of lightweight models and global average pooling to provide a large receptive field. ResNet~\cite{he2016deep} employs the residual learning and efficient bottleneck design, whose feature map is reduced by half and the number of feature maps doubles. SwiftNet~\cite{orvsic2019defense} utilizes a lightweight encoder based on ResNet and MobileNet, and designs the lateral connection architecture to reuse features from various stages, which achieves an excellent trade-off between accuracy and speed.

\begin{figure}[t]
  \centering
  \includegraphics[scale=0.5] {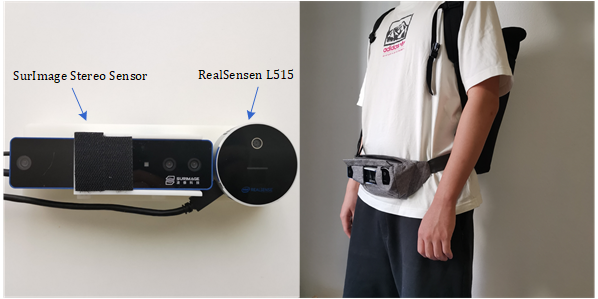}
  \caption{RealSense Camera L515 is integrated on the belt and connected with the processor NVIDIA Jetson AGX. As a multi-sensor system, SurImage Stereo Camera complements the RealSense in various conditions.}
  \label{hardware}
  \vskip-2ex
\end{figure}

\subsection{Scene Recognition}
The contributions of recent 
research works in the scene recognition area~\cite{wang2017weakly}\cite{cheng2018scene}\cite{jiang2019deep} suggest that finding potential correlations of objects in the individual scene is beneficial to classify scenes.
As the common objects in the scene may lead to misclassification, many methods ~\cite{xie2017LG}\cite{zhao2019reproducible}\cite{carlos2018in} incorporate context and explore the objects' distributions.
LG-CNN~\cite{xie2017LG} enhances fine-grained recognition by detecting local candidates and constructing a CNN architecture with local parts and global discrimination. DisNet~\cite{zhao2019reproducible} utilizes a discriminative map to select scale-aware discriminative locations for multi-scale feature extraction.
In~\cite{carlos2018in}, an extension of the DeepLab network is constructed by leveraging SVM classifiers and object histograms for scene categorization.
VASD~\cite{cheng2018scene} designs a semantic descriptor with objectness to discriminate objects and the authors observe correlations between objects among different scene classes.
However, this method, which is dependent on object information, lacks information on spatial interrelations between instances.
Scene recognition is also a topic of assistive technology for the visually impaired. For example, the unified system~\cite{cheng2020unifying} simultaneously achieves scene recognition and visual localization for people with visual impairment.

\section{Hardware System}

The unified perception framework is deployed on a portable system to support navigation assistance for the visually impaired, as it is shown in Fig.~\ref{hardware}.
It consists of a RealSense LiDAR Camera L515 and a SurImage stereo sensor mounted on a wearable belt, and an NVIDIA Jetson AGX Xavier processor that can be easily carried in the pocket or in a light-weight backpack.
The RealSense camera uses a solid-state LiDAR depth technology, which enables power-efficient (less than 3.5W) and high-quality 3D information streaming.
The SurImage sensor uses three RGB-IR cameras to form multi-baseline depth estimation and enables naturally-aligned RGB-IR-Depth information, where the stereo matching is running on its embedded FPGA processor.
In this work, we mainly use the RGB and depth images captured by the RealSense camera. Yet, as a multi-sensor system, the infrared and long-range depth information obtained by the SurImage camera can support perception in various conditions.

\begin{figure}[t]
  \centering
  \includegraphics[scale=0.75] {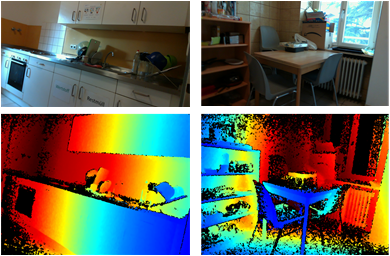}
  \caption{Top: examples of L515 color stream; Bottom: corresponding examples of L515 depth stream. The resolution of depth image is 640$\times$480, while the resolution of captured RGB image is 1280$\times$720. In practice, we align the depth image with the color image.}
  \label{realsense}
  \vskip-2ex
\end{figure}

Fig.~\ref{realsense} displays the examples of the color stream and the depth stream of Intel RealSense L515. 
The RGB images are given to the inference part of the multi-task model. The maximum distance of objects' depth information is 9 meters. Our system intends to tell users the fronting object and scene information acoustically, where only the inferences of semantic segmentation task within 2 meters are accounted as actually nearly-fronting object for feedback.

\section{Proposed Architecture}

\subsection{Approach Overview}

\begin{figure*}
  \includegraphics[width=\textwidth,height=5 cm]{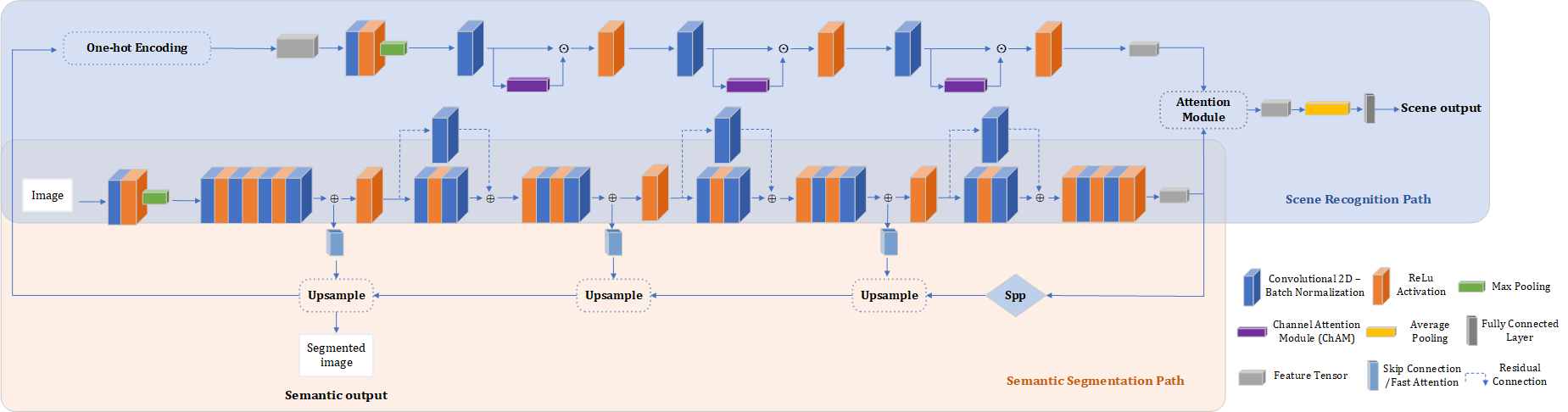}
  \caption{Our architecture based on SASceneNet~\cite{lopez2020semantic} consists of two paths: semantic segmentation path and scene recognition path. The scene recognition path composes of the upper branch responsible for extracting semantic features and the backbone with ResNet-18 for obtaining feature representations from the RGB input image. The lower path containing the backbone and upsamples is accountable for the semantic segmentation task.}
  \label{architecture}
  \vskip-2ex
\end{figure*}

We conduct a multi-task joint model composed of a semantic segmentation path and a scene classification path. They share the parameters of the backbone (we use ResNet-18~\cite{he2016deep} as illustrated in Fig.~\ref{architecture}). This recognition encoder consists of four encoder blocks corresponding to ResNet-18, and each block possesses two convolutional layers.

\subsection{Semantic Segmentation Path} 

For the semantic segmentation task, we use the single-scale architecture of SwiftNet which is an effective solution for the real-time operation on embedded devices.
The backbone ResNet18 encodes the input image and extracts the features through four convolutional groups.
The first block in the encoder produces feature maps at the H/4$\times$W/4 resolution. The following blocks downsample features by a factor of 2. At the end of the encoder, features with resolution H/32$\times$W/32 are forwarded to a simplified SPP block~\cite{orvsic2019defense} that aims to enlarge the receptive field with varying detail levels.
The upsampling decoder receives intermediate representations from the recognition encoder and passes it to upsample blocks through bilinear interpolation.
Moreover, the output of the element-wise sum within the last residual block at subsampling levels is fused to the corresponding upsampling levels in the decoder, as illustrated in Fig.~\ref{architecture}.
We integrated a Fast Attention Module~\cite{hu2020real} in the lateral connection to enhance contextual information rather than directly using skip connection. These lateral features from each stage are then aggregated and blended to the upsampling layers. 

\subsubsection{Fast Attention Module}

To increase the effectiveness of the model and avoid high computational complexity, we opt for an efficient fast attention module.
It is able to capture the contextual information across the full-resolution image by summing up the weighted features. Similar to the self-attention mechanism, the fast attention module calculates a $Value$ map containing semantic information for each pixel.
Meanwhile, it has an Affinity computation for a $Query$ map and a $Key$ map to focus on the relations between pixel locations. Unlike the original self-attention mechanism utilizing the Softmax function, the fast attention module applies a normalized cosine similarity.

The fast attention mechanism is achieved as follows:
\begin{equation}
Y = \frac{1}{height \times weight}\hat{Q}\cdot(\hat{K}^\top \cdot V),
\end{equation}
where $\hat{K}$ and $\hat{Q}$ denote L2-normalized Key and Query, and $\hat{K}$ is first multiplied with Value $V$. The $height \times weight$ refers to the spatial size of corresponding feature map.
As its complexity is quadric with respect to the channel size instead of the spatial size, the fast attention module largely reduces the computation.

\subsection{Scene Recognition Path}

\begin{figure}[t]
  \centering
  \includegraphics[scale=0.27] {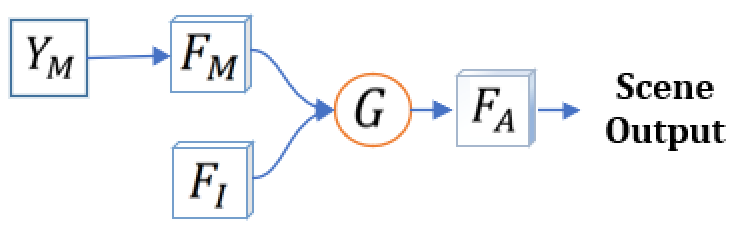}
  \caption{ The message passing chart of Scene Recognition Path. $Y_M$ indicates the logits from the semantic segmentation path, while $F_M$ and $F_I$ indicate the features produced by the 3-layer semantic extractor and the features from the backbone ResNet-18. They are then gated by the attention module at the end of the scene recognition path. The generated feature $F_A$ is used to give inference of the scene posterior probabilities.}
  \label{message_passing}
  \vskip-2ex
\end{figure}

The logits output from the semantic segmentation path is transferred as One-hoe Encoding for representing the meaningful objects and their spatial relationships. The message passing flow is presented in Fig.~\ref{message_passing}.

The scene classification path consists of the shared backbone and our proposed semantic extractor, which is constructed by three convolutional layers with channel attention modules~\cite{woo2018cbam}, so as to extract the representative features from the semantic segmentation score map. The channel attention modules reinforce the feature channels for each pixel independently based on the per-channel attention maps. Each channel attention map is used to weight the input feature maps by Hadamard product.

\subsubsection{Channel Attention Module}
The integrated channel attention concerns the channel-wise weights over each pixel independently, which is beneficial for capturing the relevant semantic information in each pixel based on corresponding probability of the semantic class.
Distinct from the fast attention module that involves contextual information over the spatial domain, the channel attention module enhances the relevant features along the channel dimension.

\subsubsection{Gated Attention Module}

\begin{figure}[t]
  \centering
  \includegraphics[scale=0.18] {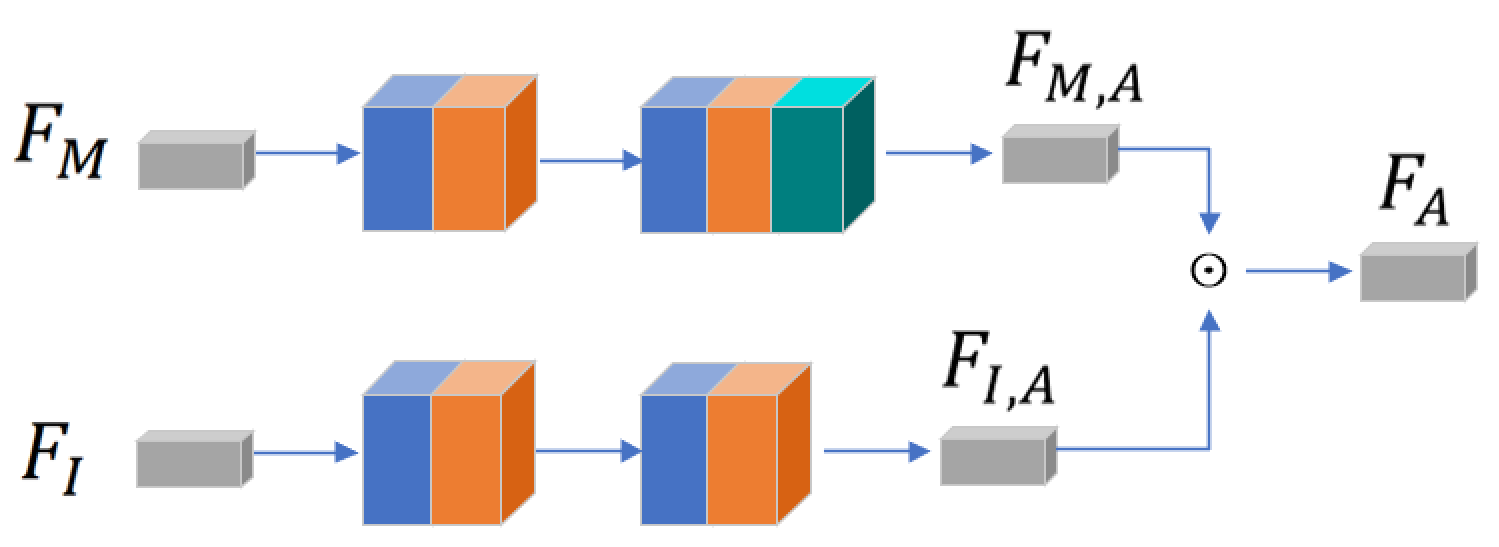}
  \caption{Each of the paths is followed by two convolutional layers. $F_M$ indicates the semantic representation produced by the semantic extracting branch. $F_I$ indicates the RGB representation produced by the backbone ResNet-18. The semantic extracting branch has an additional sigmoid layer and generates $F_{M,A}$ which is mapped to the RGB feature $F_{I,A}$ by Hadamard multiplication to gate the RGB feature. Finally, we obtain the feature $F_A$, with which the fully connected layer classifies the scene type.}
  \label{end_attention}
  \vskip-2ex
\end{figure}

Regarding texture and colour information, an attention module is adopted at the end of the scene classification path, so that it combines the RGB representation from the backbone and semantic representation from our semantic extractor. 
This attention module is shown in Fig.~\ref{end_attention}, where the RGB representation and semantic representation are individually forwarded to two convolutional layers.
After a sigmoid activation function, the semantic path obtains the normalized attention map which is multiplied to the representation from the RGB path.
This attention module ultimately leads to semantic-weighted features which are finally fed to a linear classifier. 
The scene posterior probabilities are calculated by the softmax function: 
\begin{equation}
    y_t = \log\left(\frac{\exp(f_t)}{\sum_K \exp(f_k)}\right) ,
\end{equation}
where feature vector $f$ is in the range of $\mathbb{R}^K$, $y_t$ denotes the probability for class $t$ given $f_t$.

In that way, computation of gated combination is achieved over the same numerical range, to avoid spatial information loss while scaling features in the non-gated combination.

\subsection{Joint Training and Loss}
In order to perform a joint training of multi-task model containing two different representations and to avoid the different rates of convergence, we optimize the model with a sum weighted loss. 

The Negative Log-Likelihood (NLL) loss is applied for the semantic segmentation task as shown in the following equation:
\begin{equation}
L_1 = -\log \left(y_t^{ij}\right) = -f_t^{ij} + \log\left(\sum_K \exp(f_k^{ij})\right),
\end{equation}
where $y_t^{ij}$ and $f_t^{ij}$ are the probability and feature of the $ij$-th pixel for class $t$.
The NLL loss function of the scene classification task is formulated similar to that for the semantic segmentation task:
\begin{equation}
L_2 =  -\log \left(y_t \right) = -f_t + \log\left(\sum_K \exp(f_k)\right)
\end{equation}

The final joint loss function combined by previous two losses with various weights.
\begin{equation}
\arg\min_\omega \frac{1}{N} \sum_{i=1}^N L_{final} = \arg\min_\omega \frac{1}{N} \sum_{i=1}^N \left( \lambda_1 L_1 + \lambda_2 L_2 \right),
\end{equation}
where $\lambda_1$ and $\lambda_2$ refer to weights for the semantic segmentation task and the scene recognition task. We minimize the final loss on the $N$ training samples of a given batch to find optimal parameters $\omega$ of the joint model.

\section{Experiments and Results}

\subsection{Datasets}

We trained this model on ADE20k dataset~\cite{zhou2017scene} that contains 20,210 training images fully annotated with object labels and scene labels, and 2,000 validation images. There are 150 object categories and 1,055 scene categories in the dataset.
The original images are augmented with transformation functions including random flipping, random square crop and scale, gaussian blur, and contrast augmentation, so as to increase the generalization performance. 

\subsection{Implementation Details}

The experiments are conducted with PyTorch. Each training image is cropped to 384$\times$384.
We set the initial learning rate as $ 1.0\times10^{-4}$ 
and weight decay as $ 2.5\times10^{-5}$. 
The learning rate is scheduled by Cosine Annealing. We update the weights of the network using the Adam Optimizer as the stochastic gradient descent function.

\subsection{Evaluation Metrics}

\subsubsection{Metrics for Semantic Segmentation Task}
To find the ratio of pixels properly classified, we employ pixel accuracy for evaluating the segmentation model. We also compare the performance of segmentation algorithms by the standard widely-used approach, \textit{i.e.} mean Intersection over Union (mIoU).

\subsubsection{Metrics for Scene Recognition Task}

We apply Top@$k$ accuracy metric to assess the performance of the scene classification task. The Top@$k$ reflects the proportion of validation images whose ground-truth label are contained in the $k$ top-scored classes. The commonly used assessments are Top@$k$, $k=\{1, 2, 5\}$. As classes with lower probabilities are less considered in this metric, Mean Class Accuracy (MCA) is chosen to evaluate the performance on this task, which reveals the mean of Top@$1$ metrics of all scene classes.

\subsection{Results}

\begin{table*}[htbp]
\caption{The performance on semantic segmentation task and scene recognition task.}
\vskip-2ex
\label{table1}
\begin{center}
\begin{tabular}{lllllllll}
\hline
 & Config & Pre-training & mIoU & Pixel Acc & Top@1 & Top@2 & Top@5 & MCA\\
\hline
Alternate Model & Swiftnet+SASceneNet (ResNet-18) & ImageNet & 28.26 & 68.70 & 49.80 & 59.60 & 69.70 & 15.33\\
\hline
Joint Model& ResNet-18 Baseline & ImageNet & 27.53 & 68.00 & 53.65 & 64.90 & 75.00 & 18.38\\
\hline
& ResNet-18 & scratch & 24.94 & 66.78 & 53.65 & 64.60 & 74.60 & 18.42\\

& & ImageNet & 27.80 & 68.45 & 56.85 & 67.70 & 76.25 & 21.07\\

& + Fast Attention & ImageNet & \textbf{28.60} & \textbf{69.20}& 56.20 & 67.75 & 77.15 & 21.14\\

& + Lambda Layer & ImageNet & 28.14 & 68.47 & \textbf{57.25} & \textbf{68.50} & \textbf{77.25} & \textbf{21.49}\\
\hline
& ResNet-101 & scratch & 27.04 & 68.03 & 54.85 & 65.50 & 75.65 & 21.24\\

& & ImageNet & 28.93 & 69.13 & 56.30 & 68.95 & 78.35 & 21.75\\

&  + Fast Attention & ImageNet & \textbf{31.68} & \textbf{70.66} & \textbf{57.75} & \textbf{69.45} & \textbf{78.75} & \textbf{22.21}\\

& + Lambda Layer & ImageNet & 28.16 & 68.28 & 55.68 & 65.75 & 75.55 & 22.59\\
\hline

\end{tabular}
\end{center}
\vskip-2ex
\end{table*}
 
\begin{table*}[htbp]
\caption{Computation complexity of the system, measured by GFLOPs and Parameters.}
\vskip-2ex
\label{table2}
\begin{center}
\begin{tabular}{lllll}
\hline
Config & Pre-training & GFLOPs & Params (M) \\
\hline
UperNet-50~\cite{xiao2018unified} + SASceneNet (ResNet-18)~\cite{lopez2020semantic} & scratch & 230.6 & 142.0 &\\
\hline
ResNet-18 & scratch & 26.4 & 29.1 & \\
& ImageNet & 26.4 & 29.1 & \\
+ Fast Attention & ImageNet & 27.6 & 30.1 & \\
+ Lambda layer & ImageNet & 30.9 & 29.6 & \\
+ Fast Attention and Lambda Layer & ImageNet & 32.1 & 30.6 &\\
\hline
ResNet-101 & scratch & 45.1 & 67.9 & \\
& ImageNet & 45.1 & 67.9 & \\
+ Fast Attention & ImageNet & 65.3 & 83.8 & \\
+ Lambda layer & ImageNet & 49.7 & 68.4 & \\
+ Fast Attention and Lambda Layer & ImageNet & 69.9 & 84.3 & \\
\hline 
\end{tabular}
\end{center}
\vskip-2ex
\end{table*}

We store the model whose sum of the mean IoU for semantic segmentation and its mean Class Accuracy for scene classification is the best.

Table~\ref{table1} presents comparative results for the models with various configurations in terms of: only backbone pre-trained from ImageNet, only backbone trained from scratch, Fast Attention Module, and Lambda Layers~\cite{bello2021lambdanetworks}.
The latter two constructed architecture based on ResNet-18 outperform on the scene recognition task than the baseline that solely employs the same backbone directly as the scene classifier, which indicates the semantic extracting branch helps scene recognition task.
As the pre-training is carried out on the semantic segmentation networks, incorporating the pre-trained backbone brings an improvement on scene classification metrics with respect to the models trained from scratch.
Moreover, Table~\ref{table1} also suggests that the Fast Attention Module integrated into the lateral connections increases the semantic segmentation performance.
For a comparison experimentation, we replace the channel attention layers in the semantic representation extracting branch with Lambda Layers~\cite{bello2021lambdanetworks}. It can be seen that the semantic extracting branch with lambda layers further brings slight, yet consistent enhancements on the scene classification task.
 
\begin{figure}[t!]
  \centering
  \includegraphics[scale=0.28] {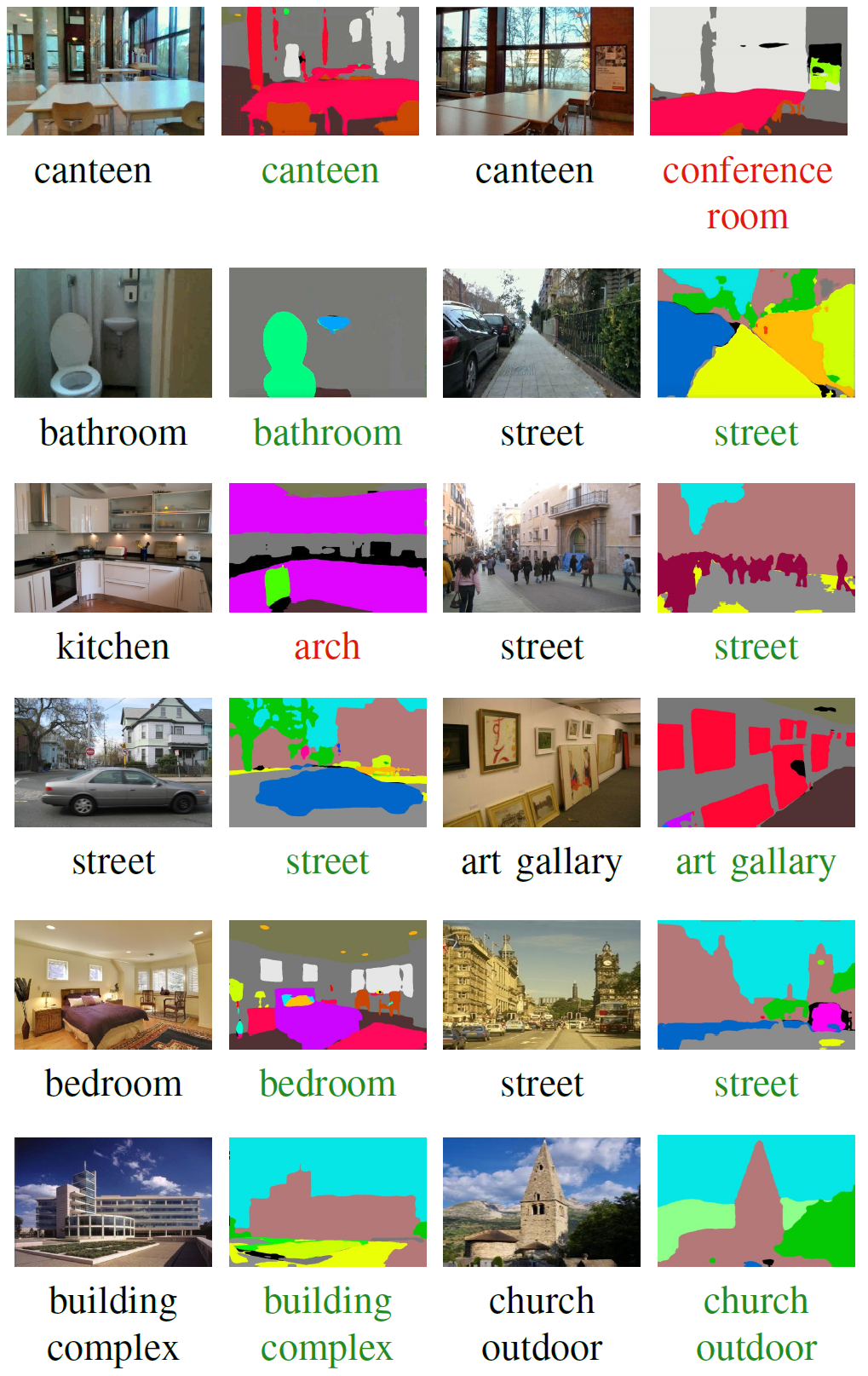}
  \caption{The first two rows show predictions of our system operated in real-time scenarios; the next rows are predictions on the ADE20k dataset. Green denotes the correct result of scene recognition; red denotes the incorrect result of scene recognition.}
  \label{inferences}
  \vskip-2ex
\end{figure}

Table \ref{table2} presents computation complexities of systems under the widely practised metric FLOPs and the number of parameters.
The accomplished model of SASceneNet~\cite{lopez2020semantic} encoding the semantic results from UperNet-50~\cite{xiao2018unified} has a drastic higher FLOPs.
Our system overall runs at 11.49 FPS, when we integrated Fast Attention and Lambda Layer configuration leveraging ResNet-18 as backbone, which is fast enough to provide navigational perception with scene recognizing and parsing predictions. 

Some inferences of our system based on ResNet-101 (the last row's configuration in Table~\ref{table1}) are presented in Fig.~\ref{inferences}. As the results of Top@$1$ presented in Fig.~\ref{inferences}, our system with Fast Attention + Lambda Layer configuration with ResNet101 as the backbone achieves satisfactory performance in real-world scenarios with a wide variety.

\begin{figure}[t]
    \centering
    \subfigure[]{
    \begin{minipage}[t]{0.19\linewidth}
    \centering
    \includegraphics[width=0.8in]{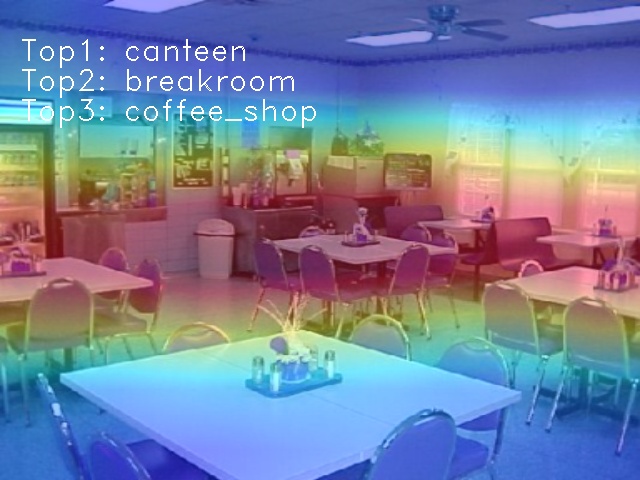}
    \end{minipage}
    }
    \subfigure[]{
    \begin{minipage}[t]{0.2\linewidth}
    \centering
    \includegraphics[width=0.8in]{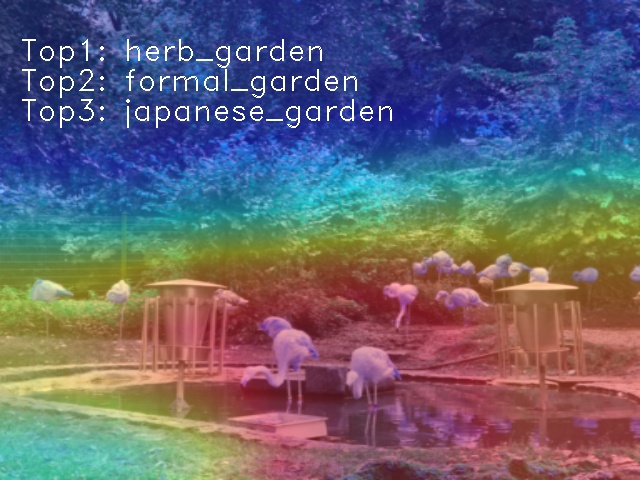}
    \end{minipage}
    }%
    \subfigure[]{
    \begin{minipage}[t]{0.2\linewidth}
    \centering
    \includegraphics[width=0.8in]{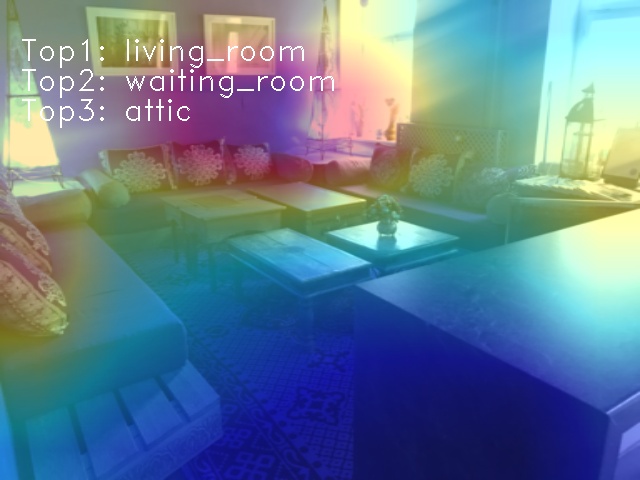}
    \end{minipage}
    }%
    \subfigure[]{
    \begin{minipage}[t]{0.2\linewidth}
    \centering
    \includegraphics[width=0.8in]{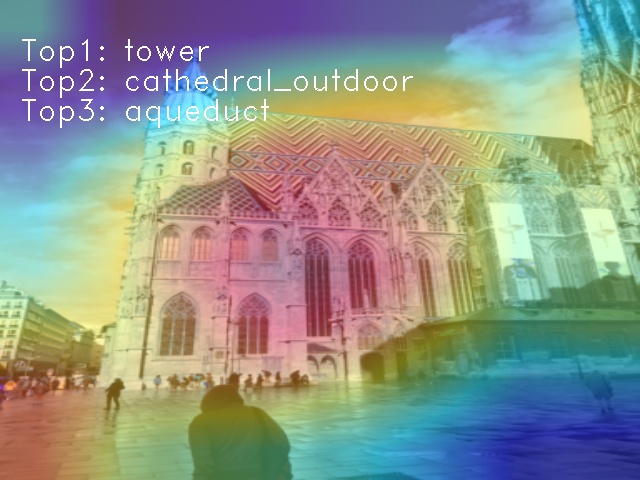}
    \end{minipage}
    }%
    \centering
    \vskip-1ex
    \caption{Class Activation Maps for scene recognition results. (a) CAM is obtained by our system on an example from ADE20k datasets. (b) to (d) CAMs are obtained by our system in real scenarios. }
    \label{cam}
    \vskip-2ex
    \end{figure}

We apply the Class Activation Map (CAM)~\cite{zhou2015cnnlocalization} to visualize which region of the input image our network focuses on for the scene recognition task. These CAMs in Fig.~\ref{cam} indicate that scene recognition highly depends on semantic-based object learning. Thereby, both quantitative and qualitative results demonstrate the effectiveness of our unified perception framework.

\section{Conclusions}

In this paper, we have designed a unified perception framework for assisting visually impaired people.
Specifically, we have presented a multi-task architecture
to build our real-time wearable system, which is able to feedback the object information and scene classes via speech signals.
The network architecture takes the feature sharing policy and 
the joint learning strategy to simultaneously fulfill both tasks.
The experiments demonstrate the effectiveness of the multi-task models with various attention modules, on both public datasets and real-world scenes captured by our portable system in indoor and outdoor environments.
Moving forward, we would like to explore the shallow architecture for more efficient multi-task learning.

\bibliographystyle{IEEEtran}
\bibliography{bib.bib}

\end{document}